\documentclass[10pt,twocolumn,letterpaper]{article}

\usepackage{iccv}
\usepackage{times}
\usepackage{epsfig}
\usepackage{graphicx}
\usepackage{amsmath}
\usepackage{amssymb}
\usepackage{makecell}
\usepackage{multirow}
\usepackage{multicol}
\usepackage{booktabs}
\usepackage{algorithm,algpseudocode,float}

\usepackage{tipa}
\usepackage[accsupp]{axessibility}
\usepackage[pagebackref=true,breaklinks=true,letterpaper=true,colorlinks,bookmarks=false]{hyperref}

\iccvfinalcopy % *** Uncomment this line for the final submission

\def\iccvPaperID{****} % *** Enter the ICCV Paper ID here

\ificcvfinal\fi

\begin{document}

%%%%%%%%% TITLE
\title{Multi-modality Associative Bridging through Memory:\\ Speech Sound Recollected from Face Video}

\author{Minsu Kim\thanks{Both authors have contributed equally to this work.} \quad\quad Joanna Hong\footnotemark[1] \quad\quad Se Jin Park \quad\quad Yong Man Ro\thanks{Corresponding author} \\
Image and Video Systems Lab, KAIST, South Korea\\
{\tt\small \{ms.k, joanna2587, jinny960812, ymro\}@kaist.ac.kr}\\
}
\maketitle

\ificcvfinal\thispagestyle{empty}\fi

%%%%%%%%% ABSTRACT
\begin{abstract}
In this paper, we introduce a novel audio-visual multi-modal bridging framework that can utilize both audio and visual information, even with uni-modal inputs. We exploit a memory network that stores source (i.e., visual) and target (i.e., audio) modal representations, where source modal representation is what we are given, and target modal representations are what we want to obtain from the memory network. We then construct an associative bridge between source and target memories that considers the interrelationship between the two memories. By learning the interrelationship through the associative bridge, the proposed bridging framework is able to obtain the target modal representations inside the memory network, even with the source modal input only, and it provides rich information for its downstream tasks. We apply the proposed framework to two tasks: lip reading and speech reconstruction from silent video. Through the proposed associative bridge and modality-specific memories, each task knowledge is enriched with the recalled audio context, achieving state-of-the-art performance. We also verify that the associative bridge properly relates the source and target memories.
\end{abstract}
\vspace{-0.4cm}
%%%%%%%%% BODY TEXT
\section{Introduction}
Recently, many studies are dealing with diverse information from multiple sources finding relationships among them \cite{ramachandram2017multimodalsurvey}. Especially, deep learning based multi-modal learning has drawn big attention with its powerful performance. While the classic approaches \cite{huang2012trad1, cao2015trad2, wang2009trad3, wang2012trad4} need to design each modal feature manually, using Deep Neural Networks (DNNs) has the advantage of automatically learning meaningful representation from each modality.
Many applications including action recognition \cite{gao2019findfusion, kim2021multispectraldetection}, object detection \cite{eitel2015multimodalod}, and image/text retrieval \cite{zhen2019dscmr} have shown the effectiveness of multi-modal learning through DNNs by analyzing a phenomenon in multi-view.

%------------------------------------ Figure 1
%#################################################
\begin{figure}[t]
	\begin{minipage}[b]{1.0\linewidth}
		\centering
		\centerline{\includegraphics[width=8.5cm]{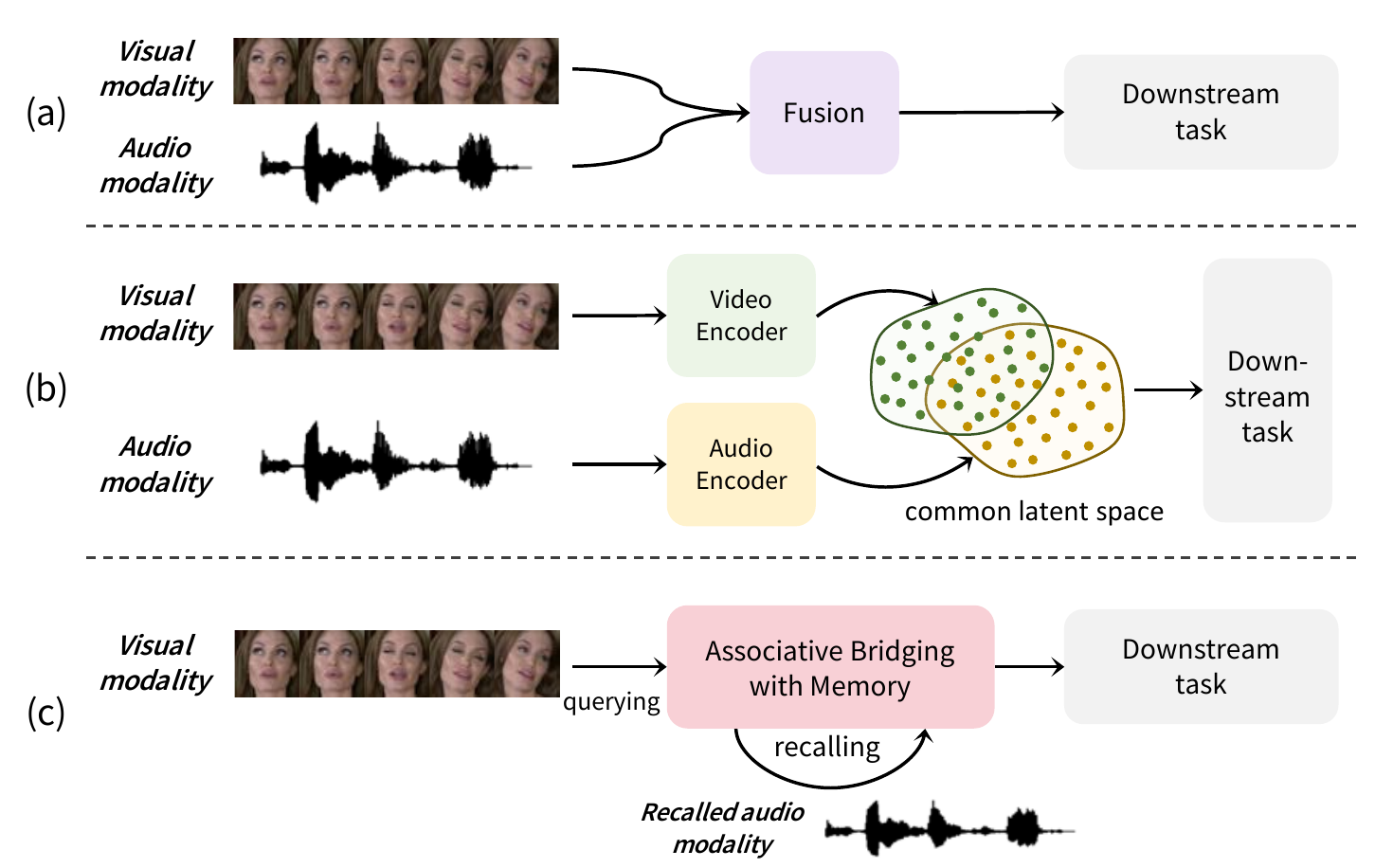}}
	\end{minipage}
	\vspace{-0.4cm}
	\caption{Illustration of audio-visual multi-modal learning. (a) Fusion of two modalities.
	(b) Learning from a common latent space of two modalities.
	(c) The proposed framework provides an associative bridge between two modalities through memory. The audio (\ie, target) modality is recalled from memory by querying the visual (\ie, source) modality. Then, both the visual and the recalled audio modalities are utilized for a downstream task.}
	\label{fig:1}
	\vspace{-0.6cm}
\end{figure}
%##################################################

Audio-visual data is one of the main ingredients for multi-modal applications such as synchronization \cite{Chung16sync,chung2019perfect}, speech recognition \cite{afouras2018deep, petridis2018end}, and speech reconstruction from silent video \cite{l2w,akbari2018lip2audspec}. Along with the rapid increase of the demands for audio-visual applications, research efforts on how to efficiently handle audio-visual data have been made. There are two main streams on handling audio-visual data. First is to extract features from the two modalities and fuse them to achieve complementary effect, as shown in Fig.\ref{fig:1} (a). Such researches \cite{petridis2018end, afouras2018deep, noda2015avsr3} try to find the most suitable architecture of DNNs to fuse the modalities. Commonly used methods are early fusion, late fusion, and intermediate fusion. These fusion methods are known to be simple, yet effectively improve the performance of a given task. However, since both modalities are necessary for the fusion, these methods cannot work when one of the modalities is missing. Second is finding a common hidden representation of two modalities by training DNNs (Fig.\ref{fig:1} (b)). Different from the first method, it can utilize the shared information of both modalities from the learned cross-modal representation with uni-modal inputs. This can be achieved by finding the common latent space of different modalities using metric learning \cite{Chung16sync,chung2019perfect} or resembling the other modality which contains rich information for a given task using knowledge distillation \cite{zhao2020hearing}. However, reducing the heterogeneity gap \cite{huang2018crossmedia}, induced by inconsistent distribution of different modalities, is still considered as a challenging problem \cite{hu2019sdml,peng2019cmgan}.

In this paper, we propose a novel multi-modal bridging framework, especially in audio speech modality and visual face modality. The proposed framework brings the advantages of the two aforementioned audio-visual multi-modal learning methods, while alleviating the problems that each method contains. That is, it can obtain both audio and visual contexts during inference even when the uni-modal input is provided only. This gives explicit complementary knowledge with the multi-modal information to uni-modal tasks which could suffer from information insufficiency. Furthermore, our work can be free from finding a common representation of different modalities, as shown in Fig.\ref{fig:1}(c).

To this end, we propose to handle the audio-visual data through memory network \cite{weston2014memory, miller2016keyvalue} which contains two modality-specific memories: source-key memory and target-value memory. Each memory stores visual and audio features arranged in pairs, respectively. Then, an associative bridge is constructed between the two modality-specific memories, to access the target-value memory by querying the source-key memory with source modal representation. Thus, when one modality (\ie, source) is given, the proposed framework can recall the other saved modality (\ie, target) from target-value memory through the associative bridge. This enables it to complement the information of uni-modal input with the recalled target modal information. Therefore, we can enrich the task-solving ability of a downstream task. The proposed framework is verified on two applications using audio-visual data: lip reading, and speech reconstruction from silent video by using visual modality as source modality and audio modality as target modality.

In summary, the major contributions of this paper are as follows:

\begin{itemize}
  \item We propose a novel audio-visual multi-modal bridging framework that enables it to utilize the information of multi-modality (\ie, audio and visual modalities) with uni-modal (\ie, visual) input during inference.
  \item We verify the effectiveness of the proposed framework on two applications: lip reading and speech reconstruction from silent video and achieve state-of-the-art performances. Moreover, we visualize that the associative bridge adequately relates the source and target memories.
  \item Through the proposed modality-specific memory operation (\ie, querying by source modality and recalling target modality), it does not need to find a common latent space of different modalities. We analyze it by comparing the proposed framework with the methods finding a common latent space of multi-modal data.
\end{itemize}

\section{Related Work}

\subsection{Multi-modal learning with audio-visual data}
Audio-visual multi-modal learning is one of the active research areas. There are two categories of audio-visual multi-modal learning using DNNs: fusion and finding a common latent space of cross-modal representation. The fusion methods \cite{tao2020avsr_multi, nefian2002avsr4, dupont2000avsr1} aim to exploit the complementary information of different modalities and achieve high performance compared to the uni-modal methods. They try to find the best fusion architecture of a given task \cite{afouras2018deep,neti2000avsr2,noda2015avsr3,lee2008avsr5}. However, as the fusion methods receive all modalities as inputs, they could not properly work if one of them is not available. The learning methods finding a common latent space from multi-modal data \cite{ngiam2011multimodal,hu2019sdml,andrew2013dcca,kan2016mvdn} aim to reduce the heterogeneity gap between the two modalities. Several works \cite{Chung16sync,chung2019perfect,feng2014crmr1} have proposed metric learning methods and adversarial learning methods to find the common representation. Other works \cite{zhao2020hearing, afouras2020asr} have proposed to learn from superior modality for a given task using knowledge distillation \cite{hinton2015distilling} which guides the learned feature to resemble the superior modal feature. 
Although finding a shared latent space or guiding one modal representation to resemble the other has the advantage of using the common information between the two modalities with uni-modal inputs, reducing the heterogeneity gap between the multi-modal data is considered as a challenging problem \cite{zhen2019dscmr,peng2019cmgan}.

In this paper, we try to not only take the advantages of both methods, but also alleviate the problems of each method. We propose to handle the audio-visual data using two modality-specific memory networks connected with an associative bridge. During inference, the proposed framework can exploit both source and the recalled target modal contexts even when the input is uni-modal. Moreover, since each modality works on its corresponding modality-specific module, we can bypass the difficulty of finding a shared latent space.

\subsection{Memory network}
Memory network is a scheme to augment neural networks using external memory \cite{weston2014memory, sukhbaatar2015endmem}. They have shown the effectiveness of memory network on modeling long-term dependencies in sequential data \cite{lee2021videomemory}. Miller \etal \cite{miller2016keyvalue} introduce key-value paired memory structure where key memory is firstly used to address relevant memories with respect to a query, extracting addressed values from the value memory. 
We utilize the key-value memory network \cite{miller2016keyvalue}, where the key memory is for saving the source modal features, and the value memory is for saving the target modal features. Thus, we can access both source and target modal contexts by recalling the saved target modal feature from the value memory when only source modality is available.

The memory network is also used in multi-modal modeling. Song \etal \cite{song2018cmmn} introduce a cross-modal memory network for cross-modal retrieval. 
Huang \etal \cite{huang2019acmm} propose an aligned cross-modal memory network for few-shot image and sentence matching. 
Using a shared memory, they encode memory-enhanced features, which will be used for image/text matching. Distinct from the previous methods, our proposed framework uses modality-specific memory network where source-key memory saves the source modality and target-value memory saves the target modality.

\subsection{Lip reading}
Lip reading is a task that recognizes speech as text from lip movements. Chung \etal \cite{chung2016lrw} propose word-level audio-visual corpus data and a baseline architecture. The performance of word-level lip reading is significantly improved with the architecture \cite{stafylakis2017reslstm, petridis2018end} of a 3D convolution layer, a ResNet-34, and Bi-RNNs. Some works \cite{weng2019twostream, xiao2020deformation} use both optical flow and video frames to capture fine-grained motion. Xu \etal \cite{xu2020discriminative} suggest a pseudo-3D CNN for the frontend which is more efficient compared to vanilla 3D CNN. Zhang \etal \cite{zhang2020cutout} show that the lip reading can be made beyond the lips by utilizing entire face as inputs. Martinez \etal \cite{martinez2020mstcn} improve the backend by changing the Bi-RNN into multi-scale temporal CNN.

It is widely known that the audio modality has superior knowledge for speech recognition than the visual modality by showing outstanding performance. In this paper, we try to complement the lip visual information by recalling the speech audio information from the proposed multi-modal bridging framework.

%------------------------------------ Figure 2
%#############################################
\begin{figure*}[t!]
	\begin{minipage}[b]{1.0\linewidth}
		\centering
		\centerline{\includegraphics[width=14.9cm]{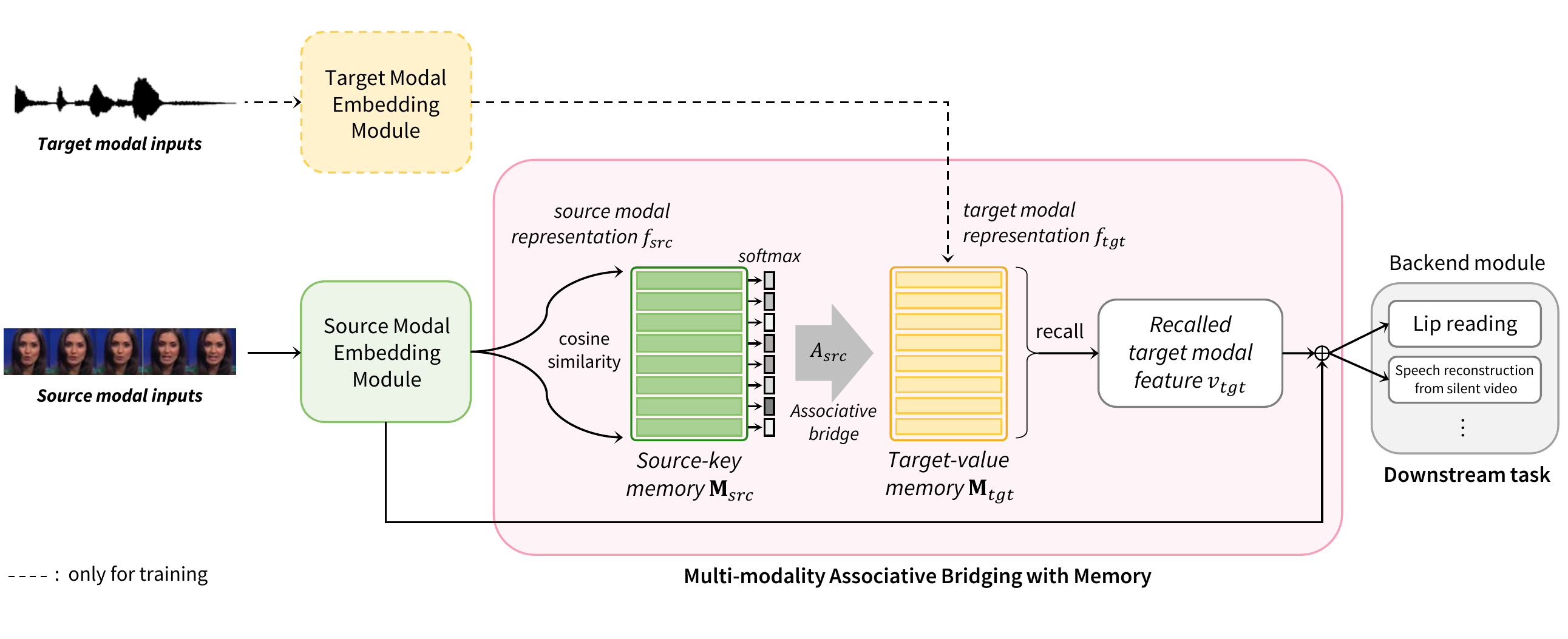}}
	\end{minipage}
	\vspace{-0.3cm}
	\caption{Overview of the proposed multi-modal bridging framework with an example of visual modality as a source and the audio modality as a target. The source-key memory is for saving source modal feature, and the target-value memory is for memorizing the target modal representations.}
	\label{fig:2}
	\vspace{-0.5cm}
\end{figure*}
%#############################################

\subsection{Speech reconstruction from silent video}
Speech reconstruction from silent video aims to generate acoustic speech signal from silent talking face video. Ephrat \etal \cite{ephrat2017vid2speech, ephrat2017improved} firstly generate speech using CNN and they improve it with a two tower CNN-based encoder-decoder architecture whose inputs are both optical flow and video frames. Akbari \etal \cite{akbari2018lip2audspec} propose to pretrain an auto-encoder to reconstruct the speech, whose decoder part is used to generate the speech from a face video. Vougioukas \etal \cite{ganbased} propose GAN based approach which maps video directly to audio waveform. Prajwal \etal \cite{l2w} attempt to learn on unconstrained single-speaker dataset. They present a model that consists of stacked 3D convolutions and an attention-based speech decoder, formulating the task as a sequence-to-sequence problem.

The speech reconstruction from silent video is considered as a challenging problem, due to the information insufficiency of face visual movement to fully represent the speech audio. We try to provide complementary information with the recalled audio representation through the proposed associative bridge with memory, and enhance its performance. With both the visual and the recalled audio contexts, we can generate high-quality speech in both speaker-dependent and speaker-independent settings.

\section{Multi-modality Associative Bridging}

The main objective of the proposed framework is to recall the target modal representation with only source modal inputs. To this end, (1) each modality-specific memory is guided to save the representative features of each modality, and (2) an associative bridge is constructed which enables it to recall the target modal representation by querying the source-key memory with source modal feature. As shown in Fig.\ref{fig:2}, the proposed multi-modality associative bridging framework is composed of two modality-specific memory networks: source-key memory $\mathbf{M}_{src}\in\mathbb{R}^{N \times C}$ and target-value memory $\mathbf{M}_{tgt}\in\mathbb{R}^{N \times D}$, where $N$ represents the number of memory slots, and $C$ and $D$ are the dimension of each modal feature, respectively. From the following subsections, we will describe the details of the proposed framework with examples of visual modality as source modality and the audio modality as target modality.

\subsection{Embedding modality-specific representations}
Each memory network inside the proposed framework saves generic representations of each modality. The generic visual and audio representations are produced from the respective modality-specific deep embedding modules. The visual (\ie, source modal) representation  $f_{src}\in\mathbb{R}^{T \times C}$ is extracted by using spatio-temporal CNN that captures both spatial and temporal information, and the audio (\ie, target modal) representation  $f_{tgt}\in\mathbb{R}^{T \times D}$ is embedded from 2D CNN whose input is preprocessed mel-spectrogram from raw audio signal, where $T$ represents the temporal length of each representation. Since the paired audio-video inputs are synchronized in time, the two embedding modules can be designed to output the same temporal length.

\subsection{Addressing modality-specific memory}
Based on the modality-specific representations, we firstly introduce how the source and target addressing vectors are formulated. The addressing vector refers to the guidance that determines where to assign weights on memory slots for a given query. Suppose that the source modal representation $f_{src}$ is given as a query, then the cosine similarity with source-key memory $\mathbf{M}_{src}$ is obtained,
\begin{align}
\label{eq:1}
    s^{i,j}_{src}=\frac{\textbf{M}_{src}^i \cdot f^j_{src} }{||\textbf{M}_{src}^i||_2 \cdot ||f^j_{src}||_2 },
\end{align}
where $s_{src}^{i,j}$ represents the cosine similarity between $i$-th memory slot of source-key memory and source modal feature in $j$-th temporal step.
Next, the relevance probability is obtained using Softmax function as follows,
\begin{align}
\label{eq:2}
    \alpha^{i,j}_{src}=\frac{\exp{(r \cdot s^{i,j}_{src})}}{\sum_{k=1}^N{\exp{(r \cdot s^{k,j}_{src})}}},
\end{align}
where $r$ is a scaling factor for similarity. By calculating the probability over the entire memory slot, the source addressing vector for the $j$-th temporal step $A^j_{src}=\{\alpha^{1,j}_{src},\alpha^{2,j}_{src},\dots \alpha^{N,j}_{src}\}$ can be obtained.

The same procedure is applied for target modal representation $f_{tgt}$ and target-value memory $\mathbf{M}_{tgt}$ to produce the target addressing vector, $A^j_{tgt}=\{\alpha^{1,j}_{tgt},\alpha^{2,j}_{tgt},\dots \alpha^{N,j}_{tgt}\}$ of $j$-th temporal step. The addressing vectors will be utilized in recalling the saved representations inside memory and connecting the two modality-specific memories, in the following subsections.

\subsection{Memorizing the target modal representations}
The obtained target addressing vector $A^j_{tgt}$ is to correctly match the target-value memory $\mathbf{M}_{tgt}$ for reconstructing target representation $\hat{f}^j_{tgt}$. To do so, the target-value memory $\mathbf{M}_{tgt}$ is trained to memorize the proper target modal representation ${f}^j_{tgt}$. We firstly obatin the reconstructed target representation $\hat{f}^j_{tgt}$ as follows,
\begin{align}
\label{eq:4}
    \hat{f}^j_{tgt} = A^j_{tgt} \cdot \mathbf{M}_{tgt}.
\end{align}
Then, we design the reconstruction loss function to guide the target-value memory $\mathbf{M}_{tgt}$ to save the proper representation. We minimize the Euclidean distance between the target representation and the reconstructed representation,
\begin{align}
\label{eq:5}
    \mathcal{L}_{save} = \mathbb{E}_j[||f^j_{tgt} - \hat{f}^j_{tgt}||^2_2].
\end{align}
With the saving loss, the target-value memory $\mathbf{M}_{tgt}$ saves the representative features of the target modality. Therefore, the recalled target modal representation $\hat{f}^j_{tgt}$ from target-value memory $M_{tgt}$ is able to represent the original target modal representation $f_{tgt}$.

\subsection{Bridging source and target memories}
\vspace{-0.11cm}
To recall the target modal representation from the target-value memory by using the source-key memory and source modal inputs, we construct an associative bridge between the two modality-specific memories. Specifically, the source-key memory is utilized to provide the bridge between source and target modalities in the form of the source addressing vector. That is, through the source addressing vector $A^j_{src}$, the corresponding saved target representation is recalled.
To achieve this, the source addressing vector $A^j_{src}$ is guided to match to the target addressing vector $A^j_{tgt}$ with the following bridging loss,
\begin{align}
\label{eq:6}
    \mathcal{L}_{bridge} = \mathbb{E}_j[D_{KL}(A^j_{tgt}||A^j_{src})],
\end{align}
where $D_{KL}(\cdot)$ represents Kullback–Leibler divergence \cite{kullback1951kld}.
With the bridging loss, the source-key memory saves the source modal representations in the same location, where the target-value memory saves the corresponding target modal features. Therefore, when a source modal representation is given, the source-key memory provides the location information of the corresponding saved target modal representation in the target-value memory, using the source addressing vector.

\subsection{Applying for downstream tasks}
\vspace{-0.11cm}
Through the associative bridge and the modality-specific memories, we can obtain the recalled target modal feature $v_{tgt}$ by using source addressing vector $A_{src}$ as follows,
\begin{align}
\label{eq:3}
    v^j_{tgt} = A^j_{src} \cdot \mathbf{M}_{tgt}.
\end{align}

Here, the target modal feature $v_{tgt}$ is recalled by querying the source-key memory $\mathbf{M}_{src}$ with the source modal representation $f_{src}$. Thus, we do not need the target modal inputs for recalling the target modal feature. Then, we can apply the recalled target modal feature for a downstream task in addition to the source modality, improving task performance by exploiting the complementary information.

\subsection{End-to-End training}
The proposed framework is trainable in an end-to-end manner, including the modality-specific embedding modules, memory networks, and the downstream sub-networks. To this end, the following task loss is applied,
\begin{align}
\label{eq:7}
    \mathcal{L}_{task} = g(h(f_{src}\oplus v_{tgt});y) + g(h(f_{src}\oplus f_{tgt});y),
\end{align}
where $g(\cdot)$ is a loss function corresponding to the downstream task, $h(\cdot)$ is a fusion layer such as a linear layer, $y$ represents label, and $\oplus$ represents concatenation. The first term of the loss function is related to the performance on a given task that utilizes both the source and the recalled target modalities. The second term guarantees that the target modal embedding module learns the meaningful representations that will be saved into target-value memory in an end-to-end manner.

Finally, the total loss function is defined as a sum of the all loss functions,
\begin{align}
\label{eq:8}
    \mathcal{L}_{total} = \mathcal{L}_{save} + \mathcal{L}_{bridge} + \mathcal{L}_{task}.
\end{align}
The pseudo code for training the proposed framework is shown at Algorithm \ref{alg:1}.

%------------------------------------ algorithm 1
%###################################################################################
\makeatletter
\renewcommand{\ALG@beginalgorithmic}{\small}
\algrenewcommand\alglinenumber[1]{\small #1:}
\makeatother
\begin{algorithm}[t!]
  \caption{Training algorithm of the proposed framework}
    \label{alg:1}
    \begin{algorithmic}[1]
        \State {\bf Inputs}: The training pairs of source and target modal inputs $(X_{src}, X_{tgt})$ and label $y$, where $X_{src} = \{x^{l}_{src}\}_{l=1}^L$, $X_{tgt}=\{x^{s}_{tgt}\}_{s=1}^S$. The learning rate $\eta$.
        \State {\bf Output}: The optimized parameters of the network $\Phi$
        \vspace{0.05in}
        \hrule
        \vspace{0.05in}
        \State Randomly initialize parameters of the network $\Phi$
        \For{each iteration}
            \State ${f}_{src}=\{{f}^j_{src}\}_{j=1}^T=$Source\_embed$(X_{src})$
            \State ${f}_{tgt}=\{{f}^j_{tgt}\}_{j=1}^T=$Target\_embed$(X_{tgt})$
            \For{$j = 1,2,...,T$}
                \State $A^j_{src} = $Softmax$(r\cdot$CosineSim$(\mathbf{M}_{src},f^j_{src}))$
                \State $A^j_{tgt} = $Softmax$(r\cdot$CosineSim$(\mathbf{M}_{tgt},f^j_{tgt}))$
                \State $\hat{f}^j_{tgt} = A^j_{tgt} \cdot \mathbf{M}_{tgt}$ 
                \State $v^j_{tgt} = A^j_{src} \cdot \mathbf{M}_{tgt}$
            \EndFor
        \State $\mathcal{L}_{save}=\sum_{j=1}^{T}||f^j_{tgt} - \hat{f}^j_{tgt}||^2_2$    
        \State $\mathcal{L}_{bridge}=\sum_{j=1}^{T}D_{KL}(A^j_{tgt}||A^j_{src})$  
        \State $\mathcal{L}_{task}=g(h(f_{src}\oplus v_{tgt});y) + g(h(f_{src}\oplus f_{tgt});y)$
        \State $\mathcal{L}_{tot} = \mathcal{L}_{save} / T + \mathcal{L}_{bridge} / T + \mathcal{L}_{task}$
        \State Update $\Phi \leftarrow \Phi -  \eta \nabla_{\Phi} \mathcal{L}_{tot}$
        \EndFor
    \end{algorithmic}
\end{algorithm}

%###############################################################################################

\section{Experiments}
The main strength of the proposed audio-visual bridging framework is that it is possible to use multi-modal representation even if only one modal input is available. Therefore, we can enhance the uni-modal downstream tasks by exploiting complementary information from the recalled modal features. We show the effectiveness of the proposed framework on two applications, lip reading and speech reconstruction from silent video, each of which takes visual modality as an input. Therefore, visual modality is utilized as a source modality and audio modality is used as a target modality.

\subsection{Application 1: Lip reading}
\label{sec:4.1}
Lip reading is a task that recognizes speech by solely depending on lip movements. We apply the proposed multi-modal bridging framework to the lip reading to complement the visual context by bringing superior knowledge of the audio through the associative bridge and to enhance the performance.

\vspace{-0.3cm}
\subsubsection{Dataset}
\vspace{-0.2cm}
We utilize two public benchmark databases for word-level lip reading, LRW \cite{chung2016lrw} and LRW-1000 \cite{yang2019lrw1000}. Both datasets are composed of 25 fps video and 16kHz audio.

{\bf LRW} \cite{chung2016lrw} is a large-scale word-level English audio-visual dataset. It includes 500 words with a maximum of 1,000 training videos each. 
For the preprocessing, the video is cropped into 136 $\times$ 136 size centered at the lip, resized into 112 $\times$ 112, and converted into grayscale. For the data augmentation, we use random horizontal flipping and random erasing for all frames in a video consistently. The audio is preprocessed using window size of 400, hop size of 160, and 80 mel-filter banks. Thus, the preprocessed mel-spectrogram has 100 fps with 80 spectral dimensional features. We use SGD optimizer, batch size of 320, and initial learning rate of 0.03. 

{\bf LRW-1000} \cite{yang2019lrw1000} is Mandarin words audio-visual dataset. It consists of 718,018 video samples with 1,000 word classes. 
The same preprocessing and data augmentation are applied as in LRW preprocessing except for cropping because the dataset is already cropped. Moreover, since the audio provided from the dataset is longer than the word boundary by 0.4-sec, we use the video as the same length as the audio. We use Adam \cite{kingma2014adam} optimizer, batch size of 60, and initial learning rate of 0.0001.

\vspace{-0.3cm}
\subsubsection{Architecture}
For the baseline architecture, we follow the typical architecture \cite{petridis2018end, stafylakis2017reslstm} whose visual embedding module consists of one 3D convolution layer and ResNet-18 \cite{he2016resnet}, and backend module is composed of 2 layered Bi-GRU \cite{schuster1997bidirectional}. We design the audio embedding module to output the same sequence length as that of the visual embedding module. For the task loss $g(\cdot)$, cross entropy loss is applied.
The details of the network architecture can be found in supplementary.

\vspace{-0.3cm}
\subsubsection{Results}
In order to verify the effectiveness of the proposed multi-modal bridging framework on complementing the visual modality with recalled audio modality, we compare the word-level lip reading using only visual modal inputs on benchmark datasets with the state-of-the-art methods. 
Table \ref{table:1} shows the overall lip reading performances on LRW and LRW-1000 datasets. Our proposed framework achieves the highest accuracies among the previous approaches on both datasets. Especially for LRW-1000, which is known to be a difficult dataset due to unbalanced training samples, the proposed method attains a large improvement of $5.58\%$ from the previous state-of-the-art method \cite{zhang2020cutout}. From this result, we can confirm that the proposed framework is even more effective for the difficult task with the ability of complementing the insufficient visual information with the recalled audio. Moreover, since our multi-modal associative bridging framework is not dependent on the downstream architecture, deep architecture such as temporal CNN can be adopted to the proposed method to improve word prediction performance.

We also conduct an ablation study with four different models for each language (\ie, $N$=0, 44, 88, 132 for English and $N$=0, 56, 112, 168 for Mandarin) to examine the effect of the number of memory slots. The ablation results on memory slot size are reported in supplementary material. For LRW, the best word accuracy of 85.41\% is achieved when $N$=88. The proposed framework improves the baseline with a margin of 1.27\%. For LRW-1000, the best word accuracy is 50.82\% when $N$=112 by improving the baseline performance with 5.89\%. 
The proposed framework improves the performance regardless of the number of memory slots from the baseline in both languages.

By employing the recalled audio feature as complementary information of the visual context, the proposed framework successfully refines the word prediction achieving state-of-the-art performance.

\begin{table}[]
    \renewcommand{\arraystretch}{1.2}
	\renewcommand{\tabcolsep}{7mm}
\resizebox{0.9999\linewidth}{!}{
\begin{tabular}{ccc}
\hline \hline
\textbf{Method}   & \textbf{LRW}  & \textbf{LRW-1000} \\ \hline
Yang \etal \cite{yang2019lrw1000}      & 83.0    & 38.19            \\
Multi-Grained \cite{wang2019multigrained}  & 83.3     & 36.91        \\
PCPG \cite{luo2020pcpg} & 83.5      & 38.70       \\
Deformation Flow \cite{xiao2020deformation}    & 84.1   & 41.93          \\
MI Maximization \cite{zhao2020mi}   & 84.4      & 38.79       \\
Face Cutout \cite{zhang2020cutout}   & 85.0     & 45.24        \\
MS-TCN \cite{martinez2020mstcn}   & 85.3     & 41.40        \\ \hline
\textbf{Proposed Method}          & \textbf{85.4}      & \textbf{50.82}           \\ \hline \hline
\end{tabular}}
    \vspace{0cm}
    \caption{Lip reading word accuracy comparison with visual modal inputs on LRW and LRW-1000 dataset.}
    \vspace{-0.5cm}
	\label{table:1}
\end{table}

\subsection{Application 2: Speech reconstruction from silent video} 
\label{sec:4.2}

Speech reconstruction from silent video is a task of inferring the speech audio signal by watching the facial video. To demonstrate the effectiveness of the proposed multi-modal bridging framework, we apply the proposed framework to the speech reconstruction from silent video task to provide the recalled audio context in an early stage of decoding for generating a high quality speech.
\vspace{-0.2cm}
\subsubsection{Dataset}
\vspace{-0.2cm}
{\bf GRID} dataset \cite{cooke2006grid} contains short English phrases with 6 words from predefined dictionary. The video and audio are sampled with rate of 25fps and 16kHz, respectively. Following \cite{ganbased,l2w}, subjects 1, 2, 4, and 29 are taken for speaker-dependent task. For the speaker-independent setting, we follow the same split as \cite{ganbased} which uses 15 subjects for training, 5 for validation, and 5 for testing. For the preprocessing, the face is detected, cropped and resized into 96 $\times$ 96 size. The audio is preprocessed with window size of 800, hop size of 160, and 80 mel-filter banks, becoming 80-dimensional mel-spectrogram in 100 fps. We use Adam optimizer, batch size of 64, and initial learning rate of 0.001.

\vspace{-0.2cm}
\subsubsection{Architecture}
\vspace{-0.2cm}
For the baseline architecture, we follow the state-of-the-art method \cite{l2w} whose visual embedding module is composed of 3D CNN and Bi-LSTM. We adopt the backend module as the decoder part of Tacotron2 \cite{tacotron2}. We utilize the same architecture of audio embedding module as lip reading experiment except for additional one convolution layer with kernel size of 5 before the Residual block. We adopt Griffin-Lim \cite{griffinlim} algorithm for audio waveform conversion. For the task loss $g(\cdot)$, L1 distance loss is applied.
More details of the network architecture can be found in supplementary material.

%------------------------------------ Figure 3
%#############################################
\begin{figure*}[t!]
	\begin{minipage}[b]{1.0\linewidth}
		\centering
		\centerline{\includegraphics[width=16.5cm]{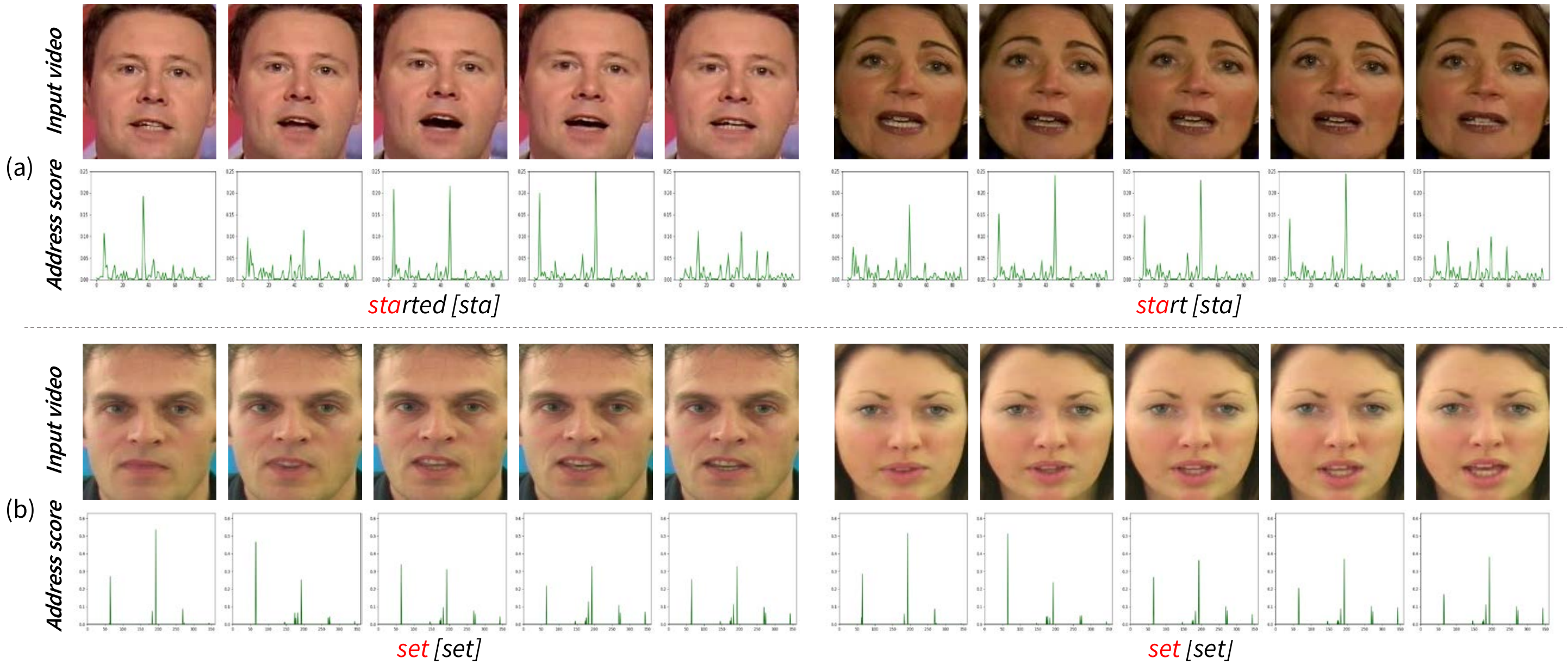}}
	\end{minipage}
	\caption{Face video clips (source modality) and corresponding addressing vectors for recalling audio modality (target modality) from learned representations inside memory: (a) results from lip reading and (b) results from speech reconstruction from silent video.}
	\label{fig:3}
	\vspace{-0.3cm}
\end{figure*}
%#############################################

\vspace{-0.2cm}
\subsubsection{Results}
\vspace{-0.2cm}
We use three standard speech quality metrics for quantitative evaluation: STOI \cite{stoi}, ESTOI \cite{estoi}, and PESQ \cite{pesq}. Table \ref{table:4} shows the performance comparison on GRID dataset in speaker-dependent setting. We report the average test scores for 4 speakers with the same setting of the previous works \cite{akbari2018lip2audspec, ganbased, ephrat2017improved, l2w, yadav2020speech}. The table clearly indicates that our model outperforms previous methods, including state-of-the-art performance. These improvements are from recalling the audio representations in the early stage of the backend which enables it to refine the generated mel-spectrogram.

\begin{table}[]
	\renewcommand{\arraystretch}{1.2}
	\renewcommand{\tabcolsep}{4mm}
\centering
\resizebox{0.89\linewidth}{!}{
\begin{tabular}{cccc}
\hline \hline
Method          & STOI & ESTOI & PESQ \\ \hline
Vid2Speech \cite{ephrat2017vid2speech}    & 0.491 & 0.335 & 1.734 \\
Lip2AudSpec \cite{akbari2018lip2audspec}   & 0.513 & 0.352 & 1.673 \\
Vougioukas \etal \cite{ganbased}     & 0.564 & 0.361 & 1.684 \\
Ephrat \etal \cite{ephrat2017improved} & 0.659 & 0.376 & 1.825 \\
Lip2Wav \cite{l2w}       & 0.731 & 0.535 & 1.772\\
Yadav \etal \cite{yadav2020speech}  & 0.724 & 0.540 & 1.932\\\hline
\textbf{Proposed Method} &  \textbf{0.738} & \textbf{0.579} & \textbf{1.984} \\ 
\hline \hline
\end{tabular}}
    \vspace{0.1cm}
    \caption{Performance of speech reconstruction comparison with visual modal inputs in a speaker-dependent setting on GRID.}
    \vspace{-0.6cm}
	\label{table:4}
\end{table}

Moreover, we ask 25 human participants to rate the Naturalness and Intelligibility. Naturalness is evaluating how natural the synthetic speech is compared to the actual human voice, and intelligibility is how clearly the words sound in the synthetic speech compared to the actual transcription. 6 samples of generated speech for each of 4 speakers of GRID are used. The human subjective evaluation results are reported at Table \ref{table:6}. Compared to the previous works \cite{ephrat2017vid2speech, l2w}, the proposed method achieves better scores on both Naturalness and Intelligibility. Moreover, with WaveNet \cite{wavenet} vocoder instead of Griffin-Lim, we can improve the scores as close to that of the ground truth. This indicates the reconstructed mel-spectrogram is of high-quality so that we can further improve the audio quality by using the state-of-the-art vocoder.

We also conduct an experiment on the speaker-independent setting, which is known to be a complex setting, of the GRID dataset to verify the effectiveness of the proposed method. As shown in Table \ref{table:5}, compared to \cite{ganbased, l2w}, the proposed framework achieves the highest performance. It can be inferred that even in a complex setting, the proposed framework can achieve meaningful outcomes by bringing the additional information through the associative bridge and memory. We visualize the examples of the generated mel-spectrogram in supplementary material.

\begin{table}[]
\centering
\renewcommand{\arraystretch}{1.2}
\renewcommand{\tabcolsep}{4.5mm}
\resizebox{0.95\linewidth}{!}{
\begin{tabular}{ccc}
\hline \hline
Method      & Naturalness  & Intelligibility  \\ \hline
Vid2Speech \cite{ephrat2017vid2speech} & 1.31 $\pm$ 0.24 & 1.42 $\pm$ 0.23 \\ 
Lip2Wav \cite{l2w} & 2.83 $\pm$ 0.21 & 2.94 $\pm$ 0.19 \\ \hline
\textbf{Proposed Method} & \textbf{2.93} $\pm$ \textbf{0.21} & \textbf{3.56} $\pm$ \textbf{0.19}  \\ \hline
\makecell{\textbf{Proposed Method}\\ (+WaveNet vocoder \cite{wavenet})} & 4.37 $\pm$ 0.16 & 4.27 $\pm$ 0.14 \\
Ground Truth & 4.62 $\pm$ 0.13 & 4.57 $\pm$ 0.14 \\ 
\hline \hline
\end{tabular}
}
\vspace{0.1cm}
\caption{Mean opinion scores for human evaluation on GRID.}
\vspace{-0.1cm}
\label{table:6}
\end{table}

\begin{table}[]
	\renewcommand{\arraystretch}{1.2}
	\renewcommand{\tabcolsep}{5mm}
\centering
\resizebox{0.9999\linewidth}{!}{
\begin{tabular}{cccc}
\hline \hline
Method          & STOI & ESTOI & PESQ \\ \hline
Vougioukas \etal \cite{ganbased}     & 0.445 & - & 1.240  \\
Lip2Wav \cite{l2w}       & 0.565 & 0.279 & 1.279\\ \hline
\textbf{Proposed Method}  &\textbf{ 0.600} & \textbf{0.315} & \textbf{1.332}\\ \hline \hline
\end{tabular}}
    \vspace{0cm}
    \caption{Performance of speech reconstruction comparison with visual modal inputs on the speaker-independent setting on GRID.}
    \vspace{-0.3cm}
	\label{table:5}
\end{table}

We conduct an ablation study on different memory slot size, which is shown in supplementary material. It shows the best scores of 0.738 STOI, 0.579 ESTOI, and 1.984 PESQ when $N$=150. Moreover, the performance of the proposed framework improves regardless of the number of memory slots, which verifies its effectiveness.

\subsection{Learned representation inside memory} 
\label{sec:4.3}
%------------------------------------ Figure 4
%#############################################
\begin{figure}[t!]
	\begin{minipage}[b]{1.0\linewidth}
		\centering
		\centerline{\includegraphics[width=7.8cm]{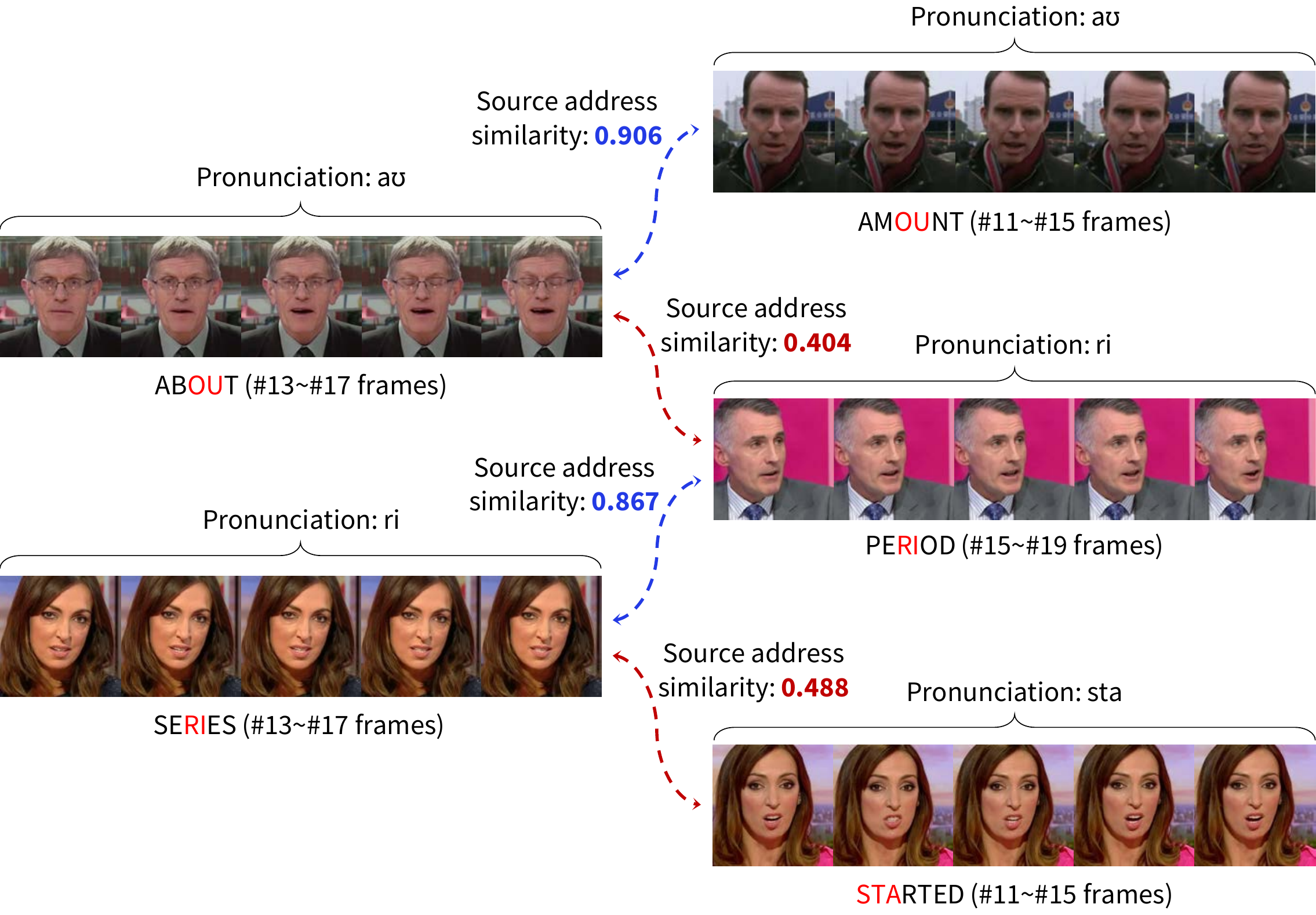}}
	\end{minipage}
	\caption{Examples of similarity between memory addressing vectors of different video clips in LRW. Note source addressing vector is for bridging video and audio modal features in memory.}
	\label{fig:4}
	\vspace{-0.5cm}
\end{figure}
%#############################################

In this section, we visualize the addressing vectors of both lip reading and speech reconstruction model in speaker-independent setting. Fig.\ref{fig:3} (a) shows the video clips of LRW dataset with consecutive 5 frames and the corresponding addressing vectors of lip reading model. From the addressing vectors of different speakers speaking the same pronunciation, we observe the similar tendency of the addressing vectors. For example, when the face video is saying ``sta'' in words \textit{started} and \textit{start}, similar memory slots are highly addressed. The same tendency can be observed in the speech reconstruction model shown in Fig.\ref{fig:3} (b). This shows that source-key memory consistently finds the corresponding saved audio location in the target-value memory by using the talking face video clips as a query, which means the associative bridge is meaningfully constructed.

In addition, we compare addressing vectors of facial video clips with different pronunciations. Fig.\ref{fig:4} shows the consecutive video frames with its corresponding pronunciation, and the comparison results. We can observe that the source addressing vectors of saying similar pronunciation have high similarity, while differently pronouncing videos have low similarity. For example, video clips of pronunciation ``a\textipa{\textupsilon}'' of word \textit{about} and \textit{amount} have 0.906 cosine similarity. In contrast, the similarity between ``ri'' of word \textit{period} and ``a\textipa{\textupsilon}'' of word \textit{about} is low with 0.404.

\subsection{Comparison with methods finding a common latent space of multi-modality} 
\label{sec:4.4}

We examine that the proposed framework can bypass the difficulty of finding a common representation of different modalities while bridging them. We compare the performance of word-level lip reading with the previous multi-modal learning methods that can exploit shared information of audio-visual modalities with uni-modal inference input by finding a common latent space. We build two multi-modal adaptation methods: cross-modal adaptation method \cite{Chung16sync} and knowledge distillation method \cite{hinton2015distilling}. The first is pretrained to synchronize the audio-visual modalities, and then trained for lip reading. The second method is additionally trained so that the features from lip reading model resemble the features from the automatic speech recognition model.

We show the word-level lip reading accuracies on LRW dataset in Table \ref{table:8}. 
By utilizing multi-modality with visual modal inputs only, all of the methods show the performance improvements from the baseline, and the proposed framework achieves the best performance. The comparison shows the efficiency of the proposed framework, where it does not need to find a common latent space of two modalities by dealing with each modality in a modality-specific memory.

%------------------------------------ Figure 5
%#############################################
\begin{figure}[t!]
	\begin{minipage}[b]{1.0\linewidth}
		\centering
		\centerline{\includegraphics[width=8cm]{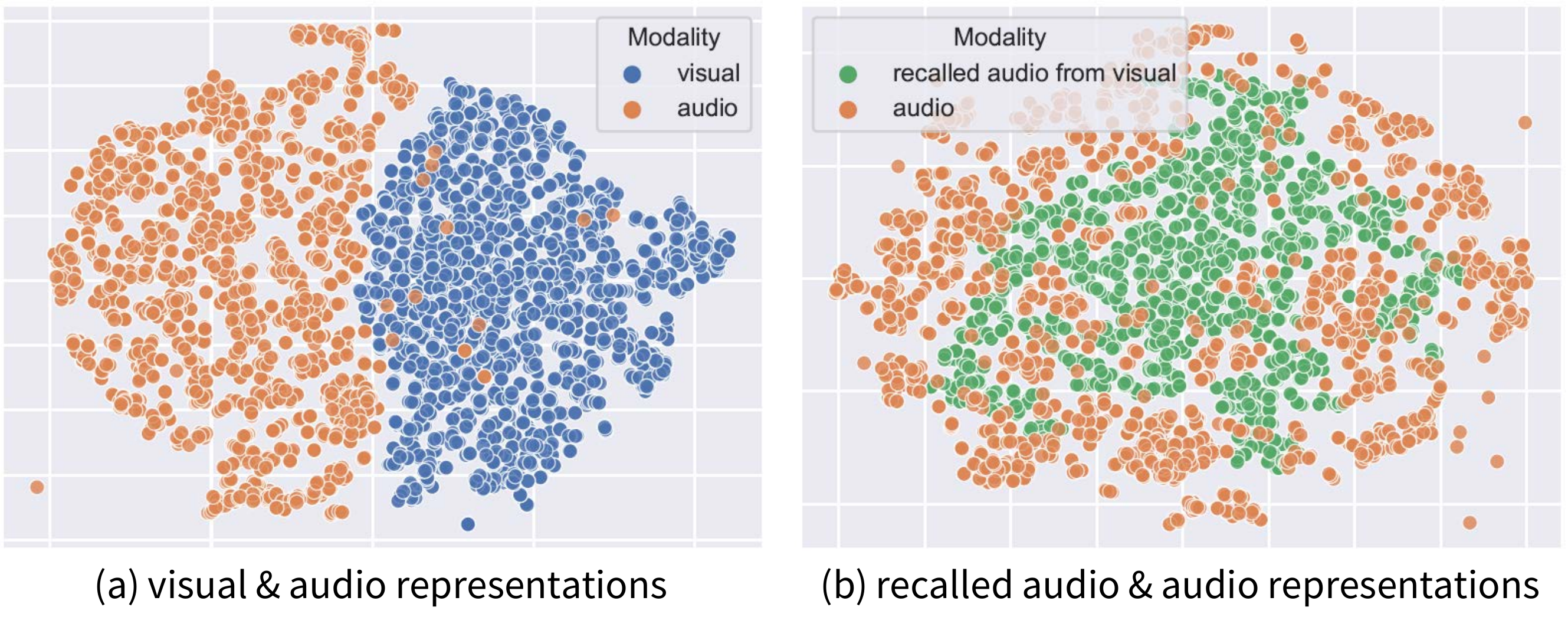}}
	\end{minipage}
	\caption{t-SNE \cite{van2008tsne} visualization of learned representation of (a) visual and audio modality, and (b) the recalled audio from visual modality and the actual audio modality.}
	\label{fig:5}
	\vspace{-0.2cm}
\end{figure}
%#############################################

\begin{table}
\centering
	\renewcommand{\tabcolsep}{1.5mm}
\resizebox{0.9999\linewidth}{!}{
\begin{tabular}{ccccc}
\hline \hline
Method & Baseline & \makecell{Cross-modal\\ Adaptation \cite{Chung16sync}} & \makecell{Knowledge \\Distillation \cite{hinton2015distilling}} & \makecell{\textbf{Proposed}\\ \textbf{Method}} \\\hline \\[-0.5em]
ACC(\%)  & 84.14 &  84.20   &  84.50  &  \textbf{85.41} \\[0.5em] \hline \hline

\end{tabular}}
    \vspace{0cm}
    \caption{Lip reading word accuracy comparison with learning methods of finding a common representation of multi-modality.}
    \vspace{-0.5cm}
	\label{table:8}
\end{table}

Lastly, we visualize the representations of visual modality, audio modality, and recalled audio modality from visual modality, by mapping them into 2D space. Fig.\ref{fig:5} shows t-SNE \cite{van2008tsne} visualization of learned representations of visual and audio modalities, and the recalled audio from visual modality and the actual audio modality. Since we handle each modality with modality-specific embedding module and memory, the two modalities have separate representations in the latent space (Fig.\ref{fig:5} (a)). However, as Fig.\ref{fig:5} (b) shows, the recalled audio from the visual modality through the associative bridge shares a representation similar to the audio modal representation. Thus, we can utilize both audio and visual contexts while maintaining their own modal representations. This visualization demonstrates that we can effectively bridging the multi-modal representations without suffering from the cross-modal adaptation by dealing with each modality in modality-specific modules.

\section{Conclusion}
In this paper, we have introduced the multi-modality associative bridging framework that connects both audio and visual context through source-key memory and target-value memory. Thus, it can utilize both audio and visual information even if only one modality is available. We have verified the effectiveness of the proposed framework on two applications: lip reading and speech reconstruction from silent video, and achieved state-of-the-art performances. Furthermore, we have shown that the proposed framework can bridge the two modalities while maintaining separate latent space for each.

{\small
\bibliographystyle{ieee_fullname}
\bibliography{egbib}
}

\end{document}